\documentclass[sigconf]{acmart}

\AtBeginDocument{%
  \providecommand\BibTeX{{%
    \normalfont B\kern-0.5em{\scshape i\kern-0.25em b}\kern-0.8em\TeX}}}

\copyrightyear{2023}
\acmYear{2023}
\setcopyright{acmcopyright}\acmConference[WSDM '23]{Proceedings of the Sixteenth ACM International Conference on Web Search and Data Mining}{February 27-March 3, 2023}{Singapore, Singapore}
\acmBooktitle{Proceedings of the Sixteenth ACM International Conference on Web Search and Data Mining (WSDM '23), February 27-March 3, 2023, Singapore, Singapore}
\acmPrice{15.00}
\acmDOI{10.1145/3539597.3570385}
\acmISBN{978-1-4503-9407-9/23/02}

\usepackage{graphicx}
\usepackage[ruled,,linesnumbered]{algorithm2e}
\usepackage{amsmath}
\usepackage{multirow}
\graphicspath{{Figures/}} 
\usepackage{amsthm}
\usepackage{stfloats}
\usepackage{balance}

\begin{document}
\title{CLNode: Curriculum Learning for Node Classification}

\author{Xiaowen Wei}
\orcid{0000-0002-7336-8434}
\affiliation{
  \department{School of Computer Science}
  \institution{Wuhan University}
  \city{Wuhan}
  \country{China}
}
\email{weixiaowen@whu.edu.cn}

\author{Xiuwen Gong}
\orcid{0000-0002-1078-1571}
\affiliation{
  \department{Faculty of Engineering}
  \institution{The University of Sydney}
  \city{Sydney}
  \country{Australia}
}
\email{xiuwen.gong@sydney.edu.au}

\author{Yibing Zhan}
\orcid{0000-0003-3180-0484}
\affiliation{
  \institution{JD Explore Academy}
  \city{Beijing}
  \country{China}
}
\email{zhanyibing@jd.com}

\author{Bo Du}
\orcid{0000-0002-0059-8458}
\affiliation{
  \department{School of Computer Science}
  \institution{Wuhan University}
  \city{Wuhan}
  \country{China}
}
\email{gunspace@163.com}

\author{Yong Luo}
\orcid{0000-0002-2296-6370}
\affiliation{
  \department{School of Computer Science}
  \institution{Wuhan University}
  \city{Wuhan}
  \country{China}
}
\email{luoyong@whu.edu.cn}

\author{Wenbin Hu}
\authornote{Corresponding author.}
\orcid{0000-0002-9258-3850}
\affiliation{
  \department{School of Computer Science}
  \institution{Wuhan University}
  \city{Wuhan}
  \country{China}
}
\email{hwb@whu.edu.cn}

\begin{abstract}
  Node classification is a fundamental graph-based task that aims to predict the classes of unlabeled nodes, for which Graph Neural Networks (GNNs) are the state-of-the-art methods. Current GNNs assume that nodes in the training set contribute equally during training. However, the quality of training nodes varies greatly, and the performance of GNNs could be harmed by two types of low-quality training nodes: (1) inter-class nodes situated near class boundaries that lack the typical characteristics of their corresponding classes. Because GNNs are data-driven approaches, training on these nodes could degrade the accuracy. (2) mislabeled nodes. In real-world graphs, nodes are often mislabeled, which can significantly degrade the robustness of GNNs. To mitigate the detrimental effect of the low-quality training nodes, we present CLNode, which employs a selective training strategy to train GNN based on the quality of nodes. Specifically, we first design a multi-perspective difficulty measurer to accurately measure the quality of training nodes. Then, based on the measured qualities, we employ a training scheduler that selects appropriate training nodes to train GNN in each epoch. To evaluate the effectiveness of CLNode, we conduct extensive experiments by incorporating it in six representative backbone GNNs. Experimental results on real-world networks demonstrate that CLNode is a general framework that can be combined with various GNNs to improve their accuracy and robustness. 
\end{abstract}



\maketitle

\section{Introduction}
Node classification is a fundamental graph-based task. Given a graph with limited labeled nodes (training nodes), the task aims to assign labels to unlabeled nodes \cite{cora-citeseer}. The state-of-the-art node classification methods are Graph Neural Networks (GNNs) \cite{GNNsurvey1,GNNsurvey2}. Generally, GNNs update the node representations by aggregating the messages passed from their neighbors. Benefiting from this aggregation mechanism, GNNs learn low-dimensional node representations that preserve the topological information and node feature attributes, which are then used to predict the labels.
Although many GNN-based node classification works \cite{GCN,graphsage,GCNII,SGC,GNNNode2} have been proposed, these works usually assume that all training nodes contribute equally. In fact, the quality of training nodes varies widely. Being data-driven approaches, GNNs exhibit degraded performance by training on the low-quality  nodes.

\begin{figure}[htb]
    \centering
    \includegraphics[width=\linewidth]{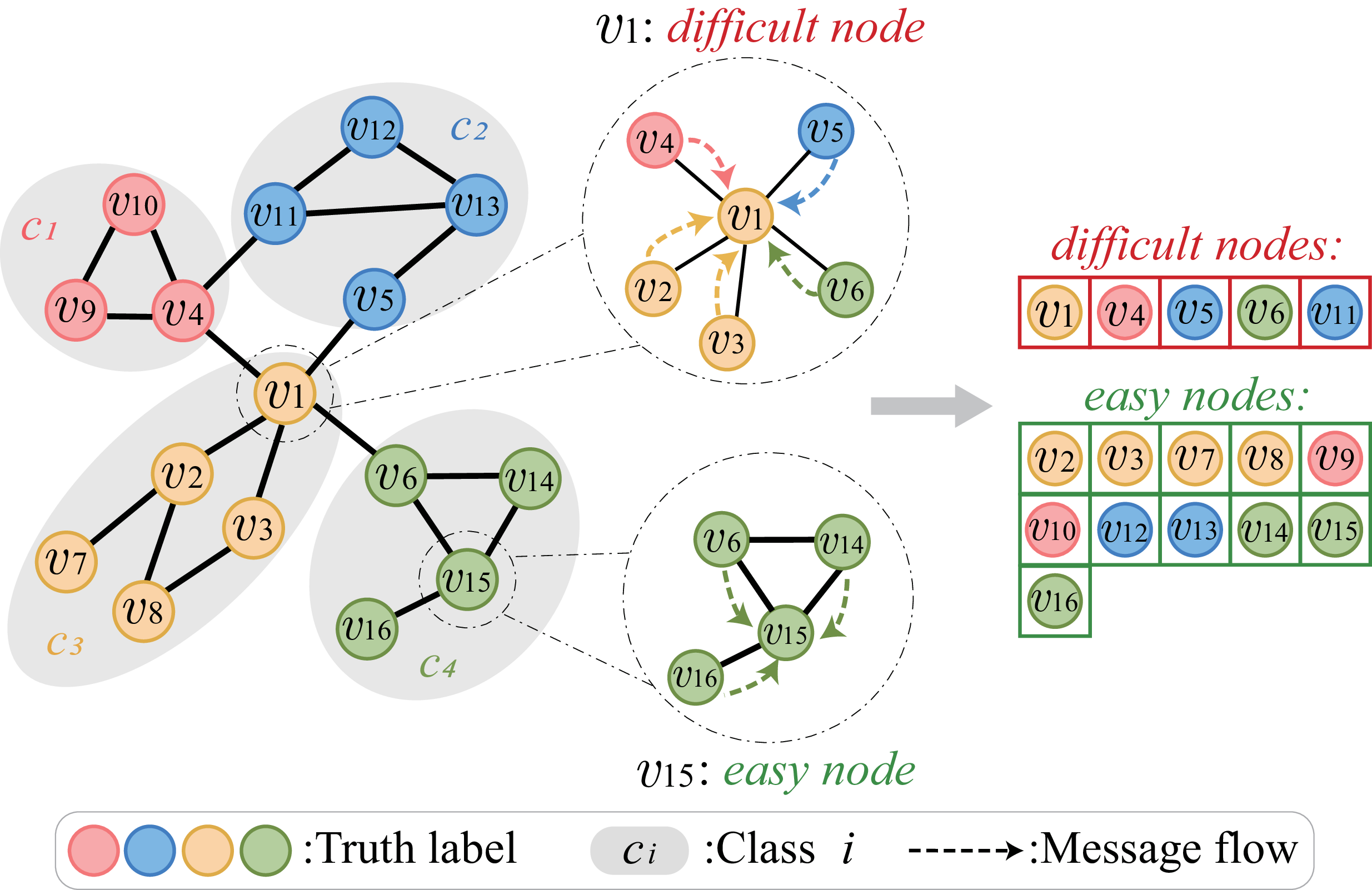}
    \caption{Illustration of node difficulty.}
    \label{intro1}
\end{figure}

To illustrate the quality of nodes, we define training nodes  whose representations lack the typical characteristics of their label classes as \textit{difficult nodes}, because it is difficult for GNNs to learn class characteristics from these low-quality nodes. In contrast, \textit{easy nodes} refer to high-quality nodes that have the typical representations of their label classes. We illustrate \textit{difficult nodes} and \textit{easy nodes} using the paper citation network in Figure \ref{intro1}. As illustrated, the cross-field paper $v_1$ connects papers from multiple classes. During neighborhood aggregation, $v_1$ aggregates messages from neighbors \{$v_2, v_3, v_4, v_5, v_6$\}. By aggregating messages \{$v_4 \rightarrow v_1, v_5 \rightarrow v_1, v_6 \rightarrow v_1$\} from classes $\{c_1, c_2, c_4\}$, $v_1$ obtains an unclear representation that mixes characteristics of different classes, indicating that $v_1$ is a \textit{difficult node}. In contrast,  all the aggregated messages of $v_{15}$ are from class $c_4$, which makes it an \textit{easy node}. Therefore, the above observation raises the question of whether these uneven-quality training nodes should be treated equally by GNNs.

\begin{figure}[htb]
    \centering
    \includegraphics[width=0.9\linewidth]{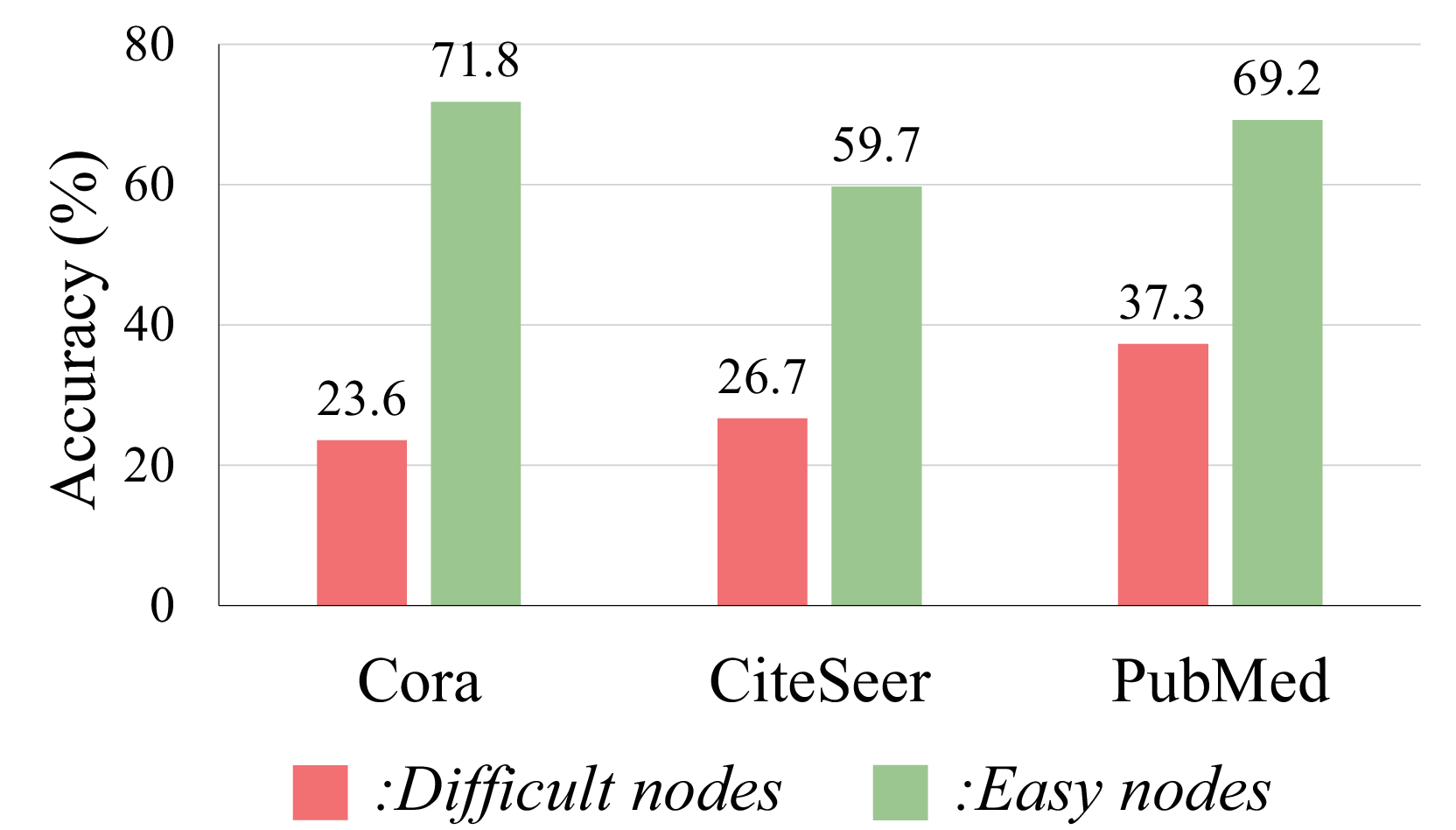}
    \caption{Accuracy of GCN trained on \textit{difficult nodes} or \textit{easy nodes}.}
    \label{intro2}
\end{figure}

\textit{Easy nodes} and \textit{difficult nodes} play different roles during training. The representations of \textit{easy nodes}  are typical, and training on such nodes helps GNNs find clear decision boundaries; whereas, \textit{difficult nodes} should be used carefully, as their representations lack the typical characteristics of their classes. There are two types of \textit{difficult nodes} that degrade the performance of GNNs: (1) inter-class nodes situated near class boundaries. By aggregating messages from neighbors, these nodes obtain unclear representations;  as a result, training on these nodes degrades the accuracy of GNNs. (2) mislabeled nodes. Real-world graphs often contain label noise\cite{dai2021nrgnn,noisylabel1,noisylabel2} and current GNNs are easily perturbed  by training on these mislabeled  nodes. 
Figure \ref{intro2} shows the accuracy of GCN \cite{GCN} on three paper citation networks \cite{cora-citeseer}, where the same number of \textit{difficult nodes} or \textit{easy nodes} are utilized for training. The node difficulty is evaluated using Eq.\eqref{eq11}, which we will detail in Section \ref{sec4}. From the results, we can see that training on \textit{easy nodes} leads to higher accuracy. For example, on the Cora network, if all training nodes are \textit{easy nodes}, the accuracy is 71.8\%, and the accuracy is only 23.6\% when only \textit{difficult nodes} are utilized.  Based on the above analysis, mitigating the detrimental effect of \textit{difficult nodes} can improve the accuracy and robustness of GNNs. In this paper, we introduce curriculum learning \cite{bengio2009CL} to mitigate the effect of these low-quality training nodes.

In particular, curriculum learning is a training strategy that initially trains the machine learning models using an easier training subset and then gradually introduces more difficult samples. By excluding low-quality difficult samples during initial training, curriculum learning mitigates overfitting to data noise, and thus improves models' accuracy and robustness \cite{CLaccuracy,CLrobust,denoise}. The most critical component of curriculum learning is the difficulty measurer, which estimates the difficulty (quality) of samples. In existing works, difficulty measurers are often designed by observing the sample features; for example, sentence length is a popular difficulty measurer in NLP tasks because shorter sentences are often easier for models to learn \cite{NLPlengthDifficultyMeasurer}. However, difficulty cannot be measured directly from node features using a similar approach. One feasible way is to utilize the graph structure, e.g., if a node connects neighbors from multiple classes, it is likely to be an inter-class \textit{difficult node}. However, in the node classification task, this is challenging due to the limited node labels.

In this paper, we attempt to address the above challenging problem by proposing a \textbf{C}urriculum \textbf{L}earning framework for \textbf{Node} Classification, called CLNode. The key idea behind CLNode is to enhance the performance of backbone GNN by incrementally introducing nodes into the training process, starting with \textit{easy nodes} and progressing to harder ones. 
Specifically, we first propose to assign pseudo-labels to unlabeled nodes. With the help of label information, we design a multi-perspective difficulty measurer, in which two difficulty measurers from local and global perspectives are proposed to measure the difficulty of training nodes.
The local difficulty measurer computes local label distribution to identify inter-class \textit{difficult nodes} because their neighbors have diverse labels; the global difficulty measurer identifies mislabeled \textit{difficult nodes} by analyzing the node feature.
Based on the measured node difficulty, we propose a continuous training scheduler that selects appropriate training nodes in each epoch to mitigate the negative effect of \textit{difficult nodes}. 
CLNode is a general framework that can be combined with various GNNs to improve their node classification performance. The key contributions of this paper are summarized as follows:
\begin{itemize}
    \item We propose CLNode, a novel curriculum learning framework for node classification. CLNode first accurately  identifies two types of \textit{difficult nodes}, and then employs a selective training strategy to mitigate the detrimental effect of these nodes.
    \item We demonstrate that CLNode can be directly plugged into existing GNNs. Without increasing the time complexity, CLNode enhances backbone GNNs by simply feeding nodes to the training process in order from easy to difficult.
    \item We conduct extensive experiments on five datasets. The results demonstrate that compared with baseline methods without curriculum learning,  CLNode effectively improves the accuracies and enhances the robustness to label noise.
\end{itemize}

\section{Related Work}
\subsection{Node Classification and GNNs}
Node classification \cite{cora-citeseer} aims to predict labels for unlabeled nodes in a given graph. As a fundamental task on graphs, node classification has various applications, including fraud detection \cite{fraud1,fraud2,fraud3}, security and privacy analytics \cite{private}, and community detection \cite{community2,community}. 

Recently, GNNs have emerged as promising approaches for analyzing graph data. Due to the long history of GNNs, we refer readers to \cite{GNNsurvey1,GNNsurvey2} for a comprehensive review. Based on the definition of graph convolution, GNNs can be broadly divided into two categories, namely spectral-based \cite{2013spectral,GCN,gat} and spatial-based \cite{graphsage,JKNET}. Bruna et al. \cite{2013spectral} first explore spectral-based GNNs by utilizing a spectral filter on the spectral space. In a follow-up work, GCN \cite{GCN} simplifies the graph convolution operation. SGC\cite{SGC} proposes to remove the nonlinearity in GCN and thereby speed up the model. Different from spectral-based methods, spatial-based methods define convolutions directly on graphs by performing operations on spatially close neighbors. GraphSAGE \cite{graphsage} is a general inductive framework that generates representations for nodes by sampling local neighbors. JK-Net \cite{JKNET} devises an alternative graph structure-based  strategy to select  neighbors for nodes. 
Although GNNs have achieved great success, they simply assume all training nodes to make equal contributions; consequently, training on the low-quality \textit{difficult nodes} significantly degrades their accuracy and robustness.

\subsection{Curriculum Learning}
Inspired by the learning principle underlying human cognitive processes, curriculum learning \cite{bengio2009CL} is proposed as a training strategy that trains machine learning models from easier samples to harder samples. Previous studies \cite{bengio2009CL,CLtheroy1,CLtheroy2} have shown that curriculum learning improves generalization capacity and guides the model towards a better parameter space.  Motivated by this, scholars have exploited the power of curriculum learning in a wide range of fields, including computer vision (CV) \cite{CLinCV,CLinCV1,CLinCV2}, natural language processing (NLP)  \cite{CLinNLP1,CLinNLP2,CLinNLP3} and graph classification \cite{curgraph}, etc. To the best of our knowledge, however, no work has yet attempted to apply curriculum learning to node classification.

\section{Preliminaries}
\subsection{Notation}
Let $\mathcal{G}=(\mathcal{V},\mathcal{E},X)$ denote a graph, where $\mathcal{V}$ is the node set, $\mathcal{E}$ is the edge set, and $X$ is the node feature matrix. The input feature of node $i$ is $x_i$, and the neighborhood of node $i$ is $\mathcal{N}(i)$. For the node classification task, a labeled node set  $\mathcal{V}_L=\{v_1,...,v_l\}$ is given with $Y_L$ denoting the input labels. $\mathcal{C}$ is the set of classes. The goal of node classification  is to predict the labels of unlabeled nodes in the graph.

\subsection{Graph Neural Networks}
Generally, a GNN involves two key computations for each node $i$ at every layer: (1) neighborhood aggregation: aggregating messages passed from $\mathcal{N}(i)$. (2) update representation: updating $i$'s representation from its representation in the previous layer and the aggregated messages. Formally, the $l$-th layer representation of node $i$ is given by:
\begin{equation}
h_i^l=\textsc{Update}(h_i^{l-1},\textsc{Aggregate}(\{h_j^{l-1} \vert j\in \mathcal{N}(i)\})).
\end{equation}

The final node representation $h_i^L$, i.e., the output of the last layer, is used for various downstream tasks. 
For the node classification task, after obtaining node representations, a multilayer perceptron is often used to map them to the predicted labels.

\subsection{Curriculum Learning}
Curriculum learning mitigates the detrimental effect of low-quality samples by using a curriculum to train the model. A curriculum is a sequence of training criteria $<Q_1, ..., Q_t,..., Q_T>$ over $T$ training epochs. Each criterion $Q_t$ is a training subset. The initial $Q_1$ consists of easier samples; as $t$ increases, more difficult samples are gradually introduced into $Q_t$. In essence, designing such a curriculum for node classification requires us to design a \textbf{difficulty measurer} and a \textbf{training scheduler}. Here, the difficulty measurer estimates the difficulty of each training node; subsequently, based on the difficulty, the training scheduler generates $Q_t$ at any training epoch $t$ to train the model. 

\begin{figure*}[t]
    \includegraphics[width=0.9\textwidth]{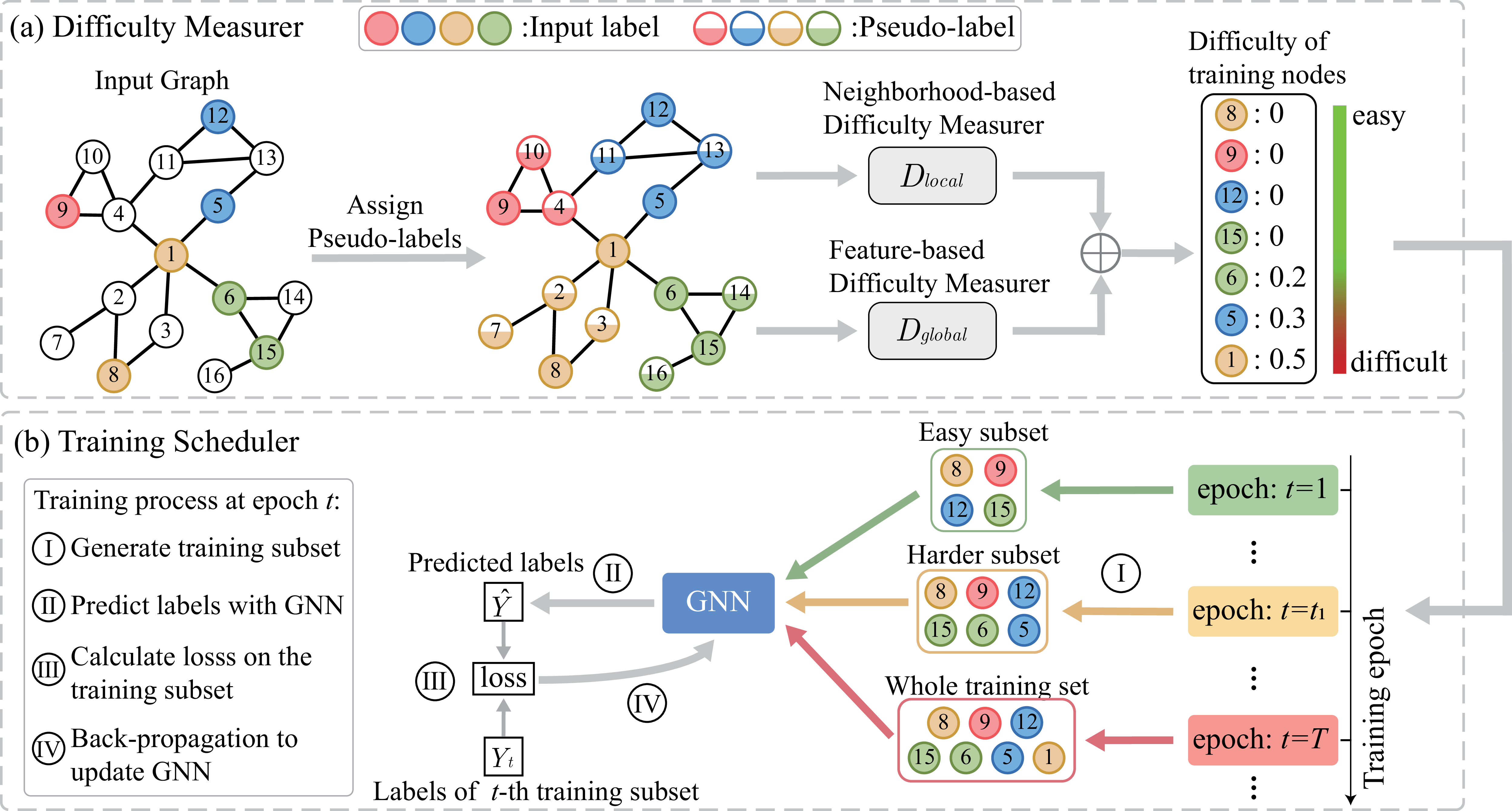}
    \caption{An overall framework of the proposed CLNode.}
    \label{framework}
\end{figure*} 

\section{Methodology}
\label{sec4}
In this section, we present the details of CLNode. As shown in Figure \ref{framework}, CLNode comprises two components: (i) multi-perspective difficulty measurer (Figure \ref{framework}(a)).  We first perform a standard node classification to obtain additional label information, then two difficulty measurers from local and global perspectives are proposed to measure the node difficulty. (ii) continuous training scheduler (Figure \ref{framework}(b)). After determining the node difficulty, we design a training scheduler to train backbone GNN with \textit{easy nodes} initially and continuously introduce harder training nodes. By paying less attention to \textit{difficult nodes}, CLNode improves the accuracy and robustness of backbone GNN. We detail the components of CLNode in the following subsections.

\subsection{Multi-perspective Difficulty Measurer}
In general, neighborhood aggregation benefits from the homophily of graphs, i.e., a node $i$'s neighbors $\mathcal{N}(i)$ tend to have the same label as $i$. 
However, the \textit{difficult nodes} violate the homophily; for example, the neighbors of an inter-class \textit{difficult node} have diverse labels because they belong to multiple classes. Taking a step further, the difficulty of nodes can be measured with the help of label information. Therefore, the first step is to assign pseudo-labels to unlabeled nodes (see Figure \ref{framework}(a)).
Specifically, we first train a GNN $f_1$ on the whole training set $\mathcal{V}_L$ to perform a standard node classification. After the training process, $f_1$ is used to get the pseudo-labels:
\begin{equation}
    H=f_1(\mathcal{G}),
    \label{eq2}
\end{equation}
\begin{equation}
    Y_{P}=MLP(H),
    \label{eq3}
\end{equation}
where $H$ is the node representation matrix obtained by GNN $f_1$ and $Y_P$ is the pseudo-labels predicted by a multilayer perceptron. However, directly using $Y_{P}$ to measure node difficulty may lead to inaccurate results, since $Y_{P}$ of training nodes may be different from the input labels $Y_{L}$. Therefore, to better measure node difficulty, we retain the input labels for training nodes:
\begin{equation}
\tilde{Y}\left[ i \right]=\begin{cases}
Y_L\left[i \right], \, i\in \mathcal{V}_L\\
Y_{P} \left[ i \right], \, otherwise.
\end{cases}
\label{eq4}
\end{equation}

Subsequently, to identify two types of \textit{difficult nodes}, i.e., inter-class nodes and mislabeled nodes, we propose two difficulty measurers to capture both local and global information for measuring the node difficulty.
\subsubsection{Neighborhood-based Difficulty Measurer}
\ \newline
We first introduce how to identify \textit{difficult nodes} from a local perspective. After obtaining $\tilde{Y}$, for each training node $u$, we calculate its difficulty with reference to the label distribution of its neighborhood. The first type of \textit{difficult nodes}  (inter-class nodes) have diverse neighbors that belong to multiple classes. In order to identify these inter-class \textit{difficult nodes}, we calculate the diversity of neighborhood's labels:

\begin{equation}
P_c(u)=\frac{\lvert\{ \tilde{Y}  \left[ v \right]=c  \, \vert \, v\in \hat{\mathcal{N}}(u)    \}\rvert}{\lvert\hat{\mathcal{N}}(u)\rvert},
\end{equation}

\begin{equation}
D_{local}(u)=-\sum_{c\in C}P_c(u)\;log(P_c(u)),
\label{eq6}
\end{equation}
where $\hat{\mathcal{N}}(u)$ denotes $\mathcal{N}(u) \cup \{u\}$ and $P_c(u)$ denotes the proportion of the neighborhood $\hat{\mathcal{N}}(u)$ belonging to class $c$. A larger  $D_{local}$ indicates a more diverse neighborhood. Taking Figure \ref{framework}(a) as an example, the $D_{local}$ of node 1 is 0.54, which is much larger than $D_{local}(8)=0$, indicating that node 1 has more diverse neighbors than node 8. Nodes with larger $D_{local}$ are more likely to be inter-class nodes. As a result, during neighborhood aggregation, these nodes aggregate neighbors’ features to get an unclear representation, making them difficult for GNNs to learn. By paying less attention to these \textit{difficult nodes},  CLNode learns more useful information and effectively improves the accuracy of backbone GNNs. 


\subsubsection{Feature-based Difficulty Measurer}
\ \newline
Because the pseudo-labels could be inaccurate, mislabeled training nodes may not be identified using local information. For instance, consider the training node 7 in Figure \ref{d-global}, whose truth label is $c_3$ but is mislabeled as $c_1$. The label information of node 7 will affect the pseudo-labels of its neighbors. As a result, the pseudo-label of node 2 is likely to be predicted as the mislabeled class $c_1$, thus the local label distribution of node 7 is consistent, from which we cannot identify it as a mislabeled node. Therefore, we propose to use global feature information to identify mislabeled nodes.

Nodes of the same class have similar features, e.g., in a paper citation network, papers in the same field tend to contain the same keywords. However, the mislabeled nodes violate this principle. For instance, in Figure \ref{d-global}, the mislabeled node 7 has low feature similarity to many nodes of its label class (e.g., node 10), since they do not in fact belong to the same class. Conversely, node 7 has high feature similarity to nodes in class $c_3$(e.g., node 8). Therefore, by exploring the feature similarity, we can deduce that node 7 is likely to be mislabeled. The input feature $X$ is sparse in high-dimensional space, instead, we use $H$ (see Eq.\eqref{eq2}) as the node feature to compute similarity. Let $h_v$ denote the feature of node $v$, then the representative feature of class $c$ is defined as the average of the node features in class $c$:
\begin{equation}
    \mathcal{V}_c = \{ v \, \vert \, \tilde{Y}[v]=c \},
\end{equation}
\begin{equation}
    h_c=\textsc{Avg}(h_v \, \vert \, v \in \mathcal{V}_c),
\end{equation}
where $\mathcal{V}_c$ denotes the nodes belonging to class $c$, and $h_c$ is the  representative feature of class $c$. To identify mislabeled \textit{difficult nodes}, for each training node $u$, we compute its feature similarity to the label class:
\begin{equation}
S(u)=\frac{exp(h_u \cdot h_{c_u})}{ \mathop{\max}_{c \in \mathcal{C}} exp(h_u \cdot h_{c})},
\label{eq9}
\end{equation}
where $c_u$ denotes the label class of node $u$, $S(u)$ calculates the feature similarity between $h_u$ and $h_{c_u}$. Mislabeled nodes tend to have smaller $S(u)$ than correctly labeled nodes.
Based on $S(u)$, the feature-based difficulty measurer is defined as:
\begin{equation}
D_{global}(u)=1-S(u).
\label{eq10}
\end{equation}

$D_{global}$ measures node difficulty from a global perspective. By using $D_{global}$ to identify mislabeled training nodes, CLNode selectively excludes these nodes from the training process, thus improving the robustness of the backbone GNNs to label noise. Considering two difficulty measurers from local and global perspectives, we finally define the difficulty of $u$ as:
\begin{equation}
\begin{aligned}
D(u)&=D_{local}(u)+ \alpha \cdot D_{global}(u),
\end{aligned}
\label{eq11}
\end{equation}
where $\alpha$ is a hyper-parameter that controls the weight of $D_{global}(u)$.

\subsection{Continuous Training Scheduler}
After measuring the node difficulty, we use a curriculum-based training strategy to train a better GNN model (see Figure \ref{framework}(b)). To distinguish it from $f_1$, we denote the model trained with curriculum as $f_2$. We propose a continuous training scheduler to generate the easy-to-difficult curriculum. In more detail, we first sort the training set $\mathcal{V}_L$ in ascending order of node difficulty; subsequently, a pacing function $g(t)$ is used to map each training epoch $t$ to a scalar $\lambda_t$ whose range is $(0,1]$, meaning that a proportion $\lambda_t$ of the easiest training nodes are used as the training subset at the $t$-th epoch. Let $\lambda_0$ denote the initial proportion of the available easiest nodes, while $T$ denotes the epoch when $g(t)$ reaches 1 for the first time. We consider three pacing functions, namely linear, root, and geometric:

\begin{figure}
    \centering
	\includegraphics[width=\linewidth]{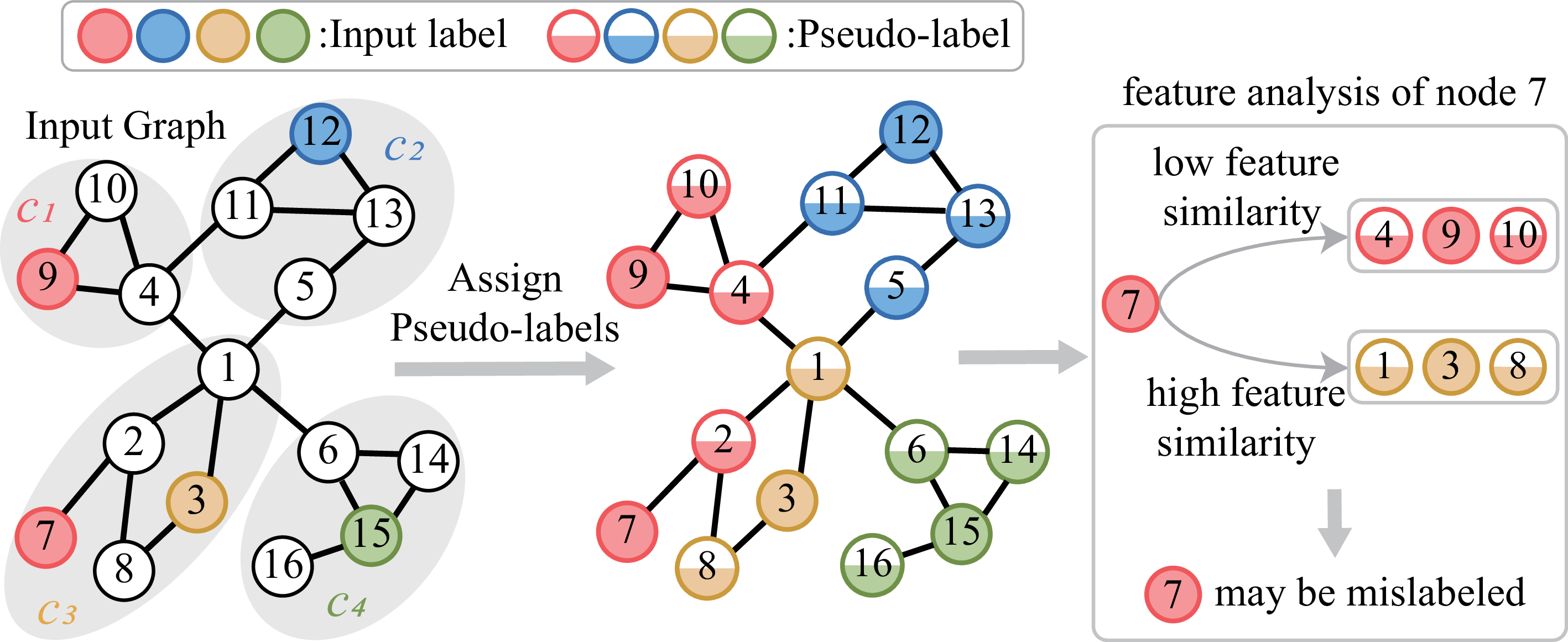}
	\caption{Illustration of the feature-based difficulty measurer.}
	\label{d-global}
\end{figure}

\begin{itemize}
    \item linear:
    \begin{equation}
        g(t)=min(1,\lambda_0+(1-\lambda_0)*\frac{t}{T}).
    \end{equation}
    \item root:
    \begin{equation}
        g(t)=min(1, \sqrt{\lambda_0^2+(1-\lambda_0^2)*\frac{t}{T}}).
    \end{equation}
    \item geometric:
    \begin{equation}
        g(t)=min(1,2^{log_2\lambda_0-log_2\lambda_0*\frac{t}{T}}).
    \end{equation}
\end{itemize}

\begin{figure}
    \centering
	\includegraphics[width=0.65\linewidth]{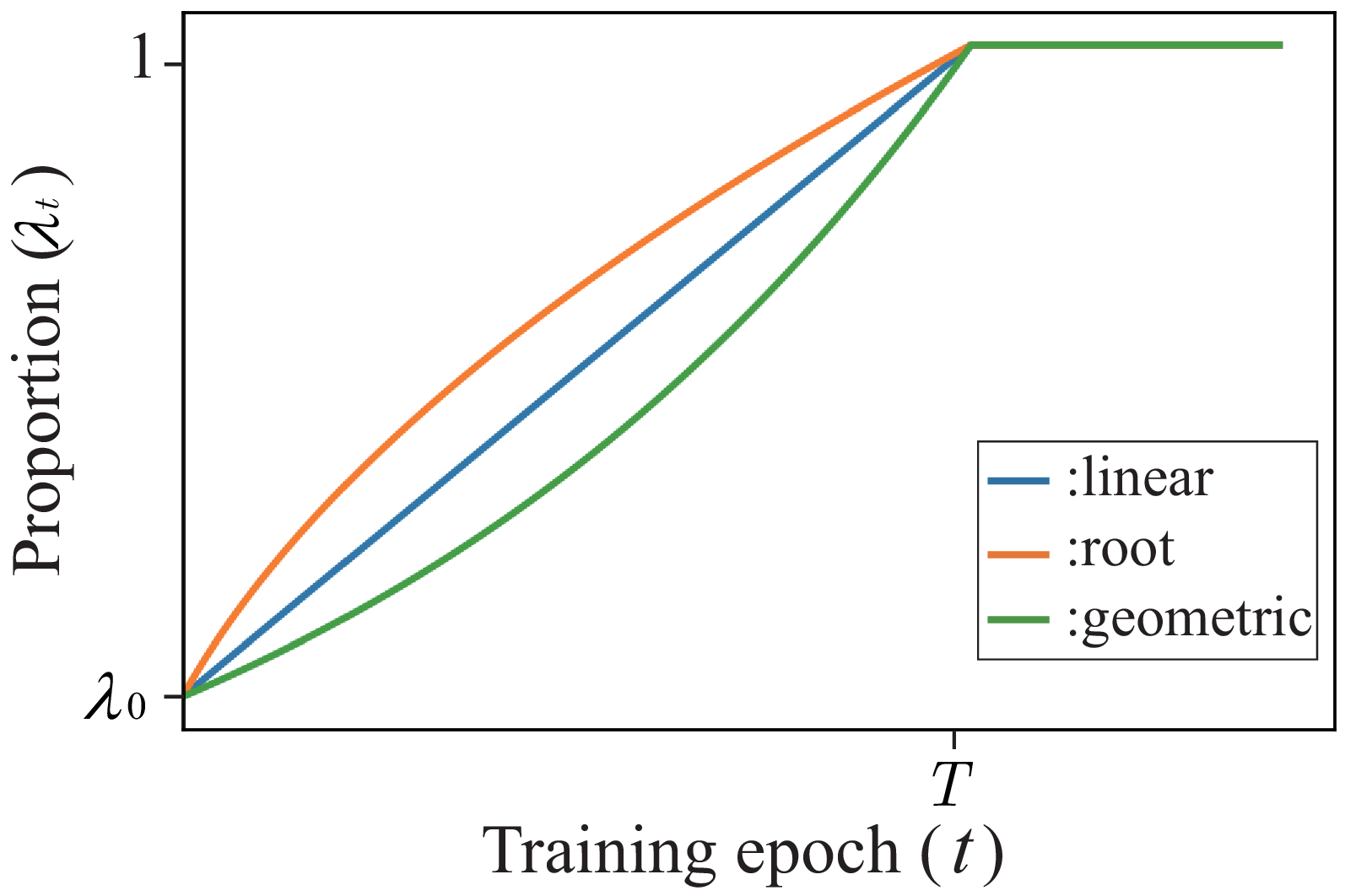}
	\caption{Visualization of three pacing functions. }
	\label{pacing}
\end{figure}

The visualization of these three pacing functions is presented in Figure \ref{pacing}. As shown in the figure, the linear function increases the difficulty of training nodes at a uniform rate; the root function introduces more \textit{difficult nodes} in fewer epochs, while the geometric function trains for a greater number of epochs on the subset of \textit{easy nodes}. By using the pacing function to continuously introduce training nodes into the training process, CLNode assigns appropriate training weights to nodes of different levels of difficulty. Specifically, the more difficult a training node is, the later it is introduced into the training process, meaning it has a smaller training weight.

Moreover, we do not stop training immediately when $t=T$, because at this time, the backbone GNN $f_2$ may not have fully explored the knowledge of nodes which have been recently introduced. Instead, when $t>T$, we use the whole training set to train $f_2$ until the test accuracy on validation set converges.


\subsection{Pseudo-code and Complexity Analysis}
\label{complexity}

In this subsection, we present the pseudo-code of CLNode and explore its time complexity. The process of CLNode is detailed in Algorithm \ref{algor1}. Lines 2--7 describe the the process of measuring node difficulty and lines 8--17 describe the process of training the backbone GNN $f_2$ with a curriculum. After the training process, $f_2$ is finally used for node classification (see line 18). As the pseudo-code shows, CLNode is easy to be plugged into any backbone GNN, as it only changes the training set in each training epoch.

\begin{algorithm}
    \caption{CLNode}
    \label{algor1}
    \KwIn{A graph $\mathcal{G}=(\mathcal{V},\mathcal{E}, X)$, the labeled node set $\mathcal{V}_L$, the input labels $Y_L$, the backbone GNN model, the hyper-parameters $\alpha$, $\lambda_0$, $T$. }
    \KwOut{The predicted labels $\hat{Y}$.}
    Initialize parameters of two GNN models $f_1$ and $f_2$\;
    Train $f_1$ on $(\mathcal{G},\mathcal{V}_L,Y_L)$\;
    Predict pseudo-labels $Y_{P}$ with $f_1$\;
    $\tilde{Y} \leftarrow$  Eq.\eqref{eq4}\;
    \For{$u \in \mathcal{V}_L$}{
        Calculate node difficulty $D(u) \gets $ Eq.\eqref{eq11}\;
    }
    Sort $\mathcal{V}_L $ according to node difficulty in ascending order\;
    Let $t=1$\;
    \While{$t<T$ or not converge}{
         $\lambda_t \gets g(t)$\;
         Generate training subset $\mathcal{V}_t \gets \mathcal{V}_L[1,...,\lfloor \lambda_t \cdot l \rfloor]$ \;
         Use $f_2$ to predict the labels $Y_t$\;
         Calculate loss $\mathcal{L}$ on $\{Y_t[v],\,Y_L[v] \, \vert \, v \in \mathcal{V}_t\}$\;
         Back-propagation on $f_2$ for minimizing $\mathcal{L}$\;
         $t \gets t+1$\;
    }
     Predict $\hat{Y}$ with $f_2$\;
\end{algorithm}

For the convenience of complexity analysis, we consider GCN as the backbone. The time complexity of an $L$-layer GCN in one epoch is $O(L\lvert \mathcal{E}\rvert F+L \lvert \mathcal{V} \rvert F^{2})$, where $F$ is the number of node feature attributes. We assume that GCN converges after $T_1$ epochs, thus its time complexity is $O(T_1\cdot(L\lvert \mathcal{E}\rvert F+L\lvert \mathcal{V} \rvert F^{2}))$, which is also the time complexity of training $f_1$. Next, the time complexity of measuring node difficulty is $O(ld+l \lvert \mathcal{C} \rvert F)$, where $d$ is the average node degree. The time complexity of sorting $\mathcal{V}_L$ is $O(l\cdot log\,l)$. Finally, we analyze the time complexity of training $f_2$. We first train $T$ epochs using the curriculum, after which we train $f_2$ with the whole $\mathcal{V}_L$ until convergence. The training of the first $T$ epochs can be seen as pre-training $f_2$ with high-quality training nodes. Therefore, $f_2$ will converge before $T+T_1$ epochs. Because $l < \lvert \mathcal{V} \rvert \ll \lvert \mathcal{E}\rvert$, the upper bound on the time complexity of CLNode is $O((2T_1+T)\cdot (L\lvert \mathcal{E}\rvert F+L\lvert \mathcal{V} \rvert F^{2}))$. In our experiments, we observe that the running time of CLNode is about twice that of the baseline GNN.

\begin{table*}[htb]
  \centering
\renewcommand\arraystretch{0.97} 
  \caption{Statistics of five benchmark datasets.}
   \begin{tabular}{lccccc}
        \toprule
        Dataset & Nodes & Edges & Features & Classes  & Label rate\\
        \midrule
        Cora  & 2708  & 5429  & 1433  & 7 & 2\%\\
        CiteSeer & 3327  & 4732  & 3703  & 6 & 2\%\\
        PubMed & 19717 & 88648 & 500  & 3 & 0.1\%\\
        A-Computers & 13381 & 245778 & 767   & 10 & 1\%\\
        A-Photo & 7487  & 119043 & 745   & 8 &1\%\\
        \bottomrule
        \end{tabular}
  \label{datasets}
\end{table*}

\begin{table*}[htb]
  \centering
  \renewcommand\arraystretch{0.98} 
  \setlength\tabcolsep{4mm}
  \caption{Node classification performance (Accuracy (\%)$\pm$Std) on five datasets.}
  \begin{tabular}{ccccccc}
    \toprule
          &   Method    & Cora  & CiteSeer & PubMed & A-Computers & A-Photo\\
    \midrule
    \multirow{3}{*}{GCN} & Original & 73.5$\pm$0.8  & 62.8$\pm$2.6   & 64.3$\pm$2.9 & 79.0$\pm$3.7 & 89.1$\pm$0.8  \\
          & +CLNode & \textbf{77.0$\pm$0.7} & \textbf{65.5$\pm$2.3} & \textbf{65.9$\pm$1.3} & \textbf{84.7$\pm$0.5} & \textbf{90.8$\pm$1.0}\\
        & (Improv.)  &3.5\%  &    2.7\% &    1.6\%     &  5.7\%   &  1.7\%  \\
    \midrule
    \multirow{3}{*}{GraphSAGE} & Original & 70.1$\pm$2.3   & 57.4$\pm$3.7 & 61.3$\pm$1.4  & 71.7$\pm$2.4  & 83.0$\pm$2.6\\
          & +CLNode & \textbf{72.1$\pm$1.4} & \textbf{60.3$\pm$3.1} & \textbf{64.1$\pm$3.8} & \textbf{77.5$\pm$1.6} & \textbf{87.5$\pm$1.2}\\
          & (Improv.)  &2.0\%  &    2.9\% &    2.8\%     &  5.8\%   &  4.5\%  \\
    \midrule
    \multirow{3}{*}{GAT} & Original & 74.2$\pm$1.2  & 63.7$\pm$2.8 & 64.6$\pm$2.5  & 80.2$\pm$0.8  & 89.4$\pm$1.8\\
          & +CLNode & \textbf{77.1$\pm$1.1} & \textbf{65.3$\pm$2.6} & \textbf{68.2$\pm$2.6} & \textbf{82.6$\pm$1.1} & \textbf{90.1$\pm$1.1}\\
          & (Improv.)  &2.9\%  &    1.6\% &    3.6\%     &  2.4\%   &  0.7\%  \\
    \midrule
    \multirow{3}{*}{SuperGAT} & Original & 74.4$\pm$4.3  & \textbf{64.8$\pm$3.3} & 67.4$\pm$4.3  & 81.2$\pm$2.0   & 87.3$\pm$2.0\\
          & +CLNode & \textbf{75.5$\pm$2.7} & 63.0$\pm$3.2 & \textbf{72.2$\pm$3.0} & \textbf{83.4$\pm$2.4} & \textbf{88.8$\pm$1.2}\\
          & (Improv.)  &1.1\%  &    - &    4.8\%     &  2.2\%   &  1.5\%  \\
    \midrule
    \multirow{3}{*}{JK-Net} & Original &74.0$\pm$1.5   & 62.1$\pm$3.7 & \textbf{66.0$\pm$1.7}  & 83.2$\pm$1.3   & 89.2$\pm$0.7\\
          & +CLNode & \textbf{76.8$\pm$0.8} & \textbf{63.6$\pm$1.2} & \textbf{71.5$\pm$3.2} & \textbf{84.4$\pm$1.0} & \textbf{90.4$\pm$0.9}\\
          & (Improv.)  &2.8\%  &    1.5\% &    5.5\%     &  1.2\%   &  1.2\%  \\
    \midrule
    \multirow{3}{*}{GCNII} & Original &76.2$\pm$4.0   & 64.5$\pm$4.3 & 70.8$\pm$6.1  & 79.8$\pm$1.8   & 87.4$\pm$2.1\\
          & +CLNode & \textbf{77.8$\pm$2.1} & \textbf{66.5$\pm$2.2} & \textbf{71.3$\pm$4.6} & \textbf{82.2$\pm$1.5} & \textbf{89.3$\pm$2.0}\\
          & (Improv.)  & 1.6\%  &  2.0\% &  0.5\% &  2.4\%   &  1.9\%  \\
    \bottomrule
    \end{tabular}
  \label{node classification}
\end{table*}

\section{Experiments}
In this section, we first evaluate the improvement in accuracy achieved by CLNode over various backbone GNNs. Further experiments are conducted on graphs with label noise to demonstrate the robustness of CLNode. Subsequently, we conduct ablation studies to verify the effectiveness of components in CLNode. Finally, we discuss the parameter sensitivity to hyper-parameters.

We conduct experiments on five benchmark datasets: Cora, Citeseer, PubMed \cite{cora-citeseer}, Amazon Computers (A-Computers), and Amazon Photo (A-Photo) \cite{coauthor-amazon}. Cora, CiteSeer, and PubMed are paper citation networks while A-Computers and A-Photo are product co-purchase networks.
Experiments are conducted on these datasets with random splits and standard splits. The random splits follow \cite{randomsplit,randomsplit2} to randomly label a specific proportion of nodes as the training set, and the label rates are listed in Table \ref{datasets}; the standard splits follow \cite{GCN,gat} in using 20 labeled nodes per class as the training set. In each dataset, we follow \cite{gat,SGC} to use 500 nodes for validation and 1000 nodes for testing.

We use six popular GNNs as the backbone models, namely GCN  \cite{GCN}, GraphSAGE \cite{graphsage}, GAT \cite{gat},  SuperGAT \cite{supergat}, JK-Net \cite{JKNET} and GCNII \cite{GCNII}, which are representative of a broad range of GNNs. In more detail, GCN is a typical convolution-based GNN,  GraphSAGE can be applied to inductive learning, GAT and SuperGAT use attention mechanism in neighborhood aggregation, while JK-Net and GCNII are deep GNNs. We use backbone GNNs without curriculum learning as baselines to explore the improvement achieved by CLNode. All models are implemented in PyTorch-geometric \cite{pyg}. We use the Adam optimizer with a learning rate of 0.01 and the weight decay is $5\times10^{-4}$.
The hidden unit is fixed at 16 in paper citation networks and 64 in product co-purchase networks. We apply two graph convolutional layers for GCN, GAT, GraphSage, and SuperGAT, 6 layers for JK-Net, and 64 layers for GCNII.
To facilitate fair comparison, the backbone GNNs' parameters of CLNode are the same as the baselines. For CLNode, $\alpha$ is fixed at 1 because we observe good performance at this value. We use the geometric pacing function by default. The hyper-parameter $\lambda_0$ is searched in the range of \{0.25, 0.5, 0.75\}, while the search space of $T$ is \{50, 100, 150\}. The code is available at \href{https://github.com/wxwmd/CLNode}{https://github.com/wxwmd/CLNode}.

\subsection{Node Classification}
In this subsection, node classification experiments are conducted on five datasets. For each baseline GNN, we compare its original accuracy with the accuracy of being plugged into the CLNode framework. We conduct each experiment for ten trials to report the average test accuracy and standard deviation.

Table \ref{node classification} reports the experimental results under random splits. The results demonstrate that CLNode can be combined with six backbone GNNs and improve their accuracy on node classification. For example, on the Cora dataset, CLNode improves the test accuracy of backbone GNNs by 3.5\% (GCN), 2.0\% (GraphSAGE), 2.9\% (GAT),  1.1\% (SuperGAT), 2.8\% (JK-Net), and 1.6\% (GCNII).
The results prove that CLNode effectively mitigates the detrimental effect of \textit{difficult nodes}, thereby enabling more useful information to be learned from uneven-quality training nodes.

\begin{table}[htb]
  \centering
  \renewcommand\arraystretch{0.9} 
  \caption{Accuracy (\%) on Cora under different label rates.}
  \begin{tabular}{clccc}
    \toprule
          & Method  & 1\%   & 2\%   & 3\%\\
    \midrule
  \multirow{2}{*}{GCN} & Original & 62.4$\pm$2.7   & 73.5$\pm$0.8 & 78.6$\pm$0.6                \\
          & +CLNode    & \textbf{66.9$\pm$1.2}  & \textbf{77.0$\pm$0.7}  &  \textbf{79.7$\pm$0.6} \\
    \midrule
    \multirow{2}{*}{GraphSage} & Original &  54.8$\pm$3.0 & 70.1$\pm$2.3 &76.0$\pm$0.8\\
          & +CLNode    & \textbf{61.8$\pm$2.6} & \textbf{72.1$\pm$1.4} &\textbf{77.7$\pm$1.5}\\
    \midrule
    \multirow{2}{*}{GAT} & Original & 65.2$\pm$2.4 & 74.2$\pm$1.2 & 78.8$\pm1.0$\\
          & +CLNode  &  \textbf{68.5$\pm$2.0} & \textbf{77.1$\pm$1.1} &\textbf{79.9$\pm$0.5}\\
    \midrule
    \multirow{2}{*}{SuperGAT} & Original & 65.5$\pm$6.0 & 74.4$\pm$4.3 & \textbf{78.7$\pm$1.6}\\
          & +CLNode    & \textbf{67.9$\pm$3.2} & \textbf{75.5$\pm$2.7} & 78.5$\pm$2.4\\
    \midrule
    \multirow{2}{*}{JK-Net} & Original &67.5$\pm$1.7 & 74.0$\pm$1.5 &77.4$\pm$1.4\\
          & +CLNode    & \textbf{69.4$\pm$1.4} & \textbf{76.8$\pm$0.8} &\textbf{78.8$\pm$0.3}\\
    
     \midrule
    \multirow{2}{*}{GCNII} & Original & 68.5$\pm$3.9 & 76.2$\pm$4.0 & 79.0$\pm$2.2\\
          & +CLNode    & \textbf{71.2$\pm$3.8} & \textbf{77.8$\pm$2.1} &\textbf{80.2$\pm$2.0}\\
    \bottomrule
    \end{tabular}
  \label{label rate}%
\end{table}%

\begin{figure}[htb]
    \centering
    \includegraphics[width=\linewidth]{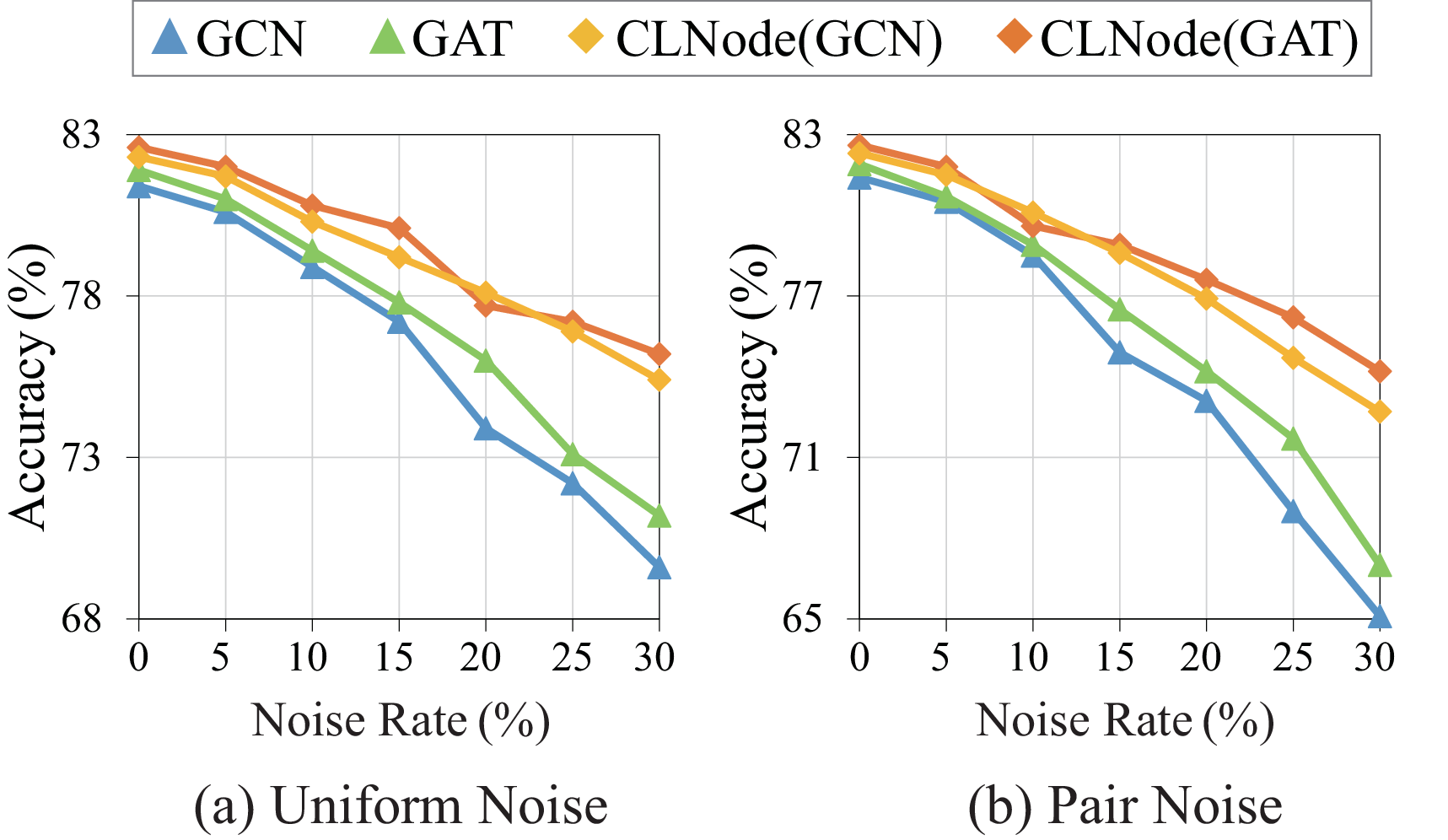}
    \caption{Accuracy (\%) on Cora with two kinds of label noise.}
    \label{noise}
\end{figure} 

Moreover, we conduct node classification experiments under different label rates. Table \ref{label rate} shows the accuracy on Cora dataset at label rates of 1\%, 2\%, 3\%, respectively. We observe that when there are fewer labeled training nodes, the improvement achieved by CLNode is more obvious. This is because when there are more training nodes, the detrimental effect of \textit{difficult nodes} is mitigated by a large number of \textit{easy nodes}; conversely, when there are fewer training nodes, \textit{difficult nodes} easily mislead GNNs to learn the wrong knowledge. Therefore, by excluding \textit{difficult nodes}
from initial training, CLNode significantly improves the accuracy of GNNs at a low label rate. For many real-world graphs, the labeling process is tedious and costly, resulting in limited labels, and it would be highly beneficial to use CLNode in these situations.

\subsection{Robustness to Noise}
In this subsection, we investigate whether CLNode enhances the robustness of backbone GNNs to label noise. In a noisily labeled graph, the labels have a probability of $p$ to be flipped to other classes, where $p$ denotes the noise rate. Following \cite{dai2021nrgnn,labelnoise}, we corrupt the labels of the training and validation set with two kinds of label noise:

\begin{itemize}
\item Uniform noise. The label has a probability of $p$ to be mislabeled as any other class.
\item Pair noise. We assume that nodes in one class can only be mislabeled as their closest class; that is, labels have a probability $p$ to flip to their pair class.
\end{itemize}

We conduct experiments on Cora under standard splits and vary $p$ from \{0, 5\%,..., 30\%\} to compare the performance of CLNode and the baseline GNNs under different levels of noise. We only report the results using GCN and GAT as backbone GNNs because we have similar observations for other GNNs. CLNode(GCN) and CLNode(GAT) denote the CLNode method using GCN and GAT as backbone GNN, respectively.


The results are shown in  Figure \ref{noise}, from which we observe that as the noise rate increases, the performance of all baselines drops dramatically. CLNode also suffers under conditions of increasing noise rate; however, when there is more noise in the graph, the performance gap between CLNode and the baseline increases. This observation demonstrates that CLNode effectively enhances the robustness of backbone GNNs to two kinds of label noise, since CLNode considers mislabeled training nodes as \textit{difficult nodes} and selectively excludes them from the training process, while the baseline GNNs treat all training nodes as equal and consequently overfit to noise.

\subsection{Ablation Study}
In this subsection, we conduct ablation studies to explore the effectiveness of the multi-perspective difficulty measurer and the sensitivity of CLNode to different pacing functions. Ablation studies are conducted on three paper citation datasets under standard splits, where the graphs are corrupted by uniform label noise and the noise rate $p$ is set to 30\%.

First, to verify the multi-perspective difficulty measurer benefits from combining the local and global information, we design two difficulty measurers to replace it:
\begin{itemize}
\item Measuring difficulty only with local information, i.e., we only use  $D_{local}$ to measure node difficulty. 
\item Measuring difficulty only with global information, i.e., we only use  $D_{global}$ to measure node difficulty.
\end{itemize}

We use these two difficulty measurers for ablation studies; in the below, we refer to the ablated methods as CLNode(local) and CLNode(global), respectively. GCN is used as the baseline method. The results are reported in Table \ref{dm-ablation}, from which we observe the following: (1) both CLNode(local) and CLNode(global) outperform the baseline method, which demonstrates that they measure the node difficulty from different perspectives, and thus mitigate the detrimental effect of different types of \textit{difficult nodes}; (2) CLNode achieves the best results in all experiments, proving that by combining local and global perspectives to measure the node difficulty, CLNode effectively identifies two types of \textit{difficult nodes}, thus enhancing the accuracy and robustness of backbone GNNs.

\begin{table}[t]
  \centering
  \caption{Comparisons between different difficulty measurers. }
  \renewcommand\arraystretch{1.1} 
  \begin{tabular}{ccccc}
    \toprule
          &   Method    & Cora  & CiteSeer & PubMed \\
    \midrule
    \multirow{4}{*}{GCN} & original & 69.6  & 55.3 & 69.4 \\
          & +CLNode(local) & 74.8 & 61.8 & 74.2\\
          & +CLNode(global) & 72.3 & 62.5 & 73.2 \\
          & +CLNode & \textbf{75.4} & \textbf{63.1} & \textbf{74.4} \\
    \bottomrule
    \end{tabular}
  \label{dm-ablation}
\end{table}

\begin{table}[]
    \centering
    \renewcommand\arraystretch{1.05} 
    \caption{Comparisons between different pacing functions.}
     \begin{tabular}{ccccc}
    \toprule
          & Pacing Function    & Cora  & CiteSeer & PubMed \\
    \midrule
          \multirow{3}{*}{CLNode} & linear & 74.8 & 62.7 & 74.2\\
          & root & 74.5 & 62.5 & 73.9 \\
          & geometric & \textbf{75.4} & \textbf{63.1} & \textbf{74.4} \\
    \bottomrule
    \end{tabular}
    \label{pacing-ablation}
\end{table}

In Table \ref{pacing-ablation}, we evaluate the sensitivity of CLNode to three pacing functions: linear, root, and geometric. We find that the geometric pacing function has a slight advantage on all datasets. As shown in Figure \ref{pacing}, the geometric function trains for a greater number of epochs on the subset of \textit{easy nodes}  before introducing \textit{difficult nodes}. Therefore, to mitigate the detrimental effect of \textit{difficult nodes}, we believe that the high-confidence knowledge in \textit{easy nodes}  should be fully explored before   more \textit{difficult nodes} are introduced.

\begin{figure}[htb]
    \centering
    \includegraphics[width=0.65\linewidth]{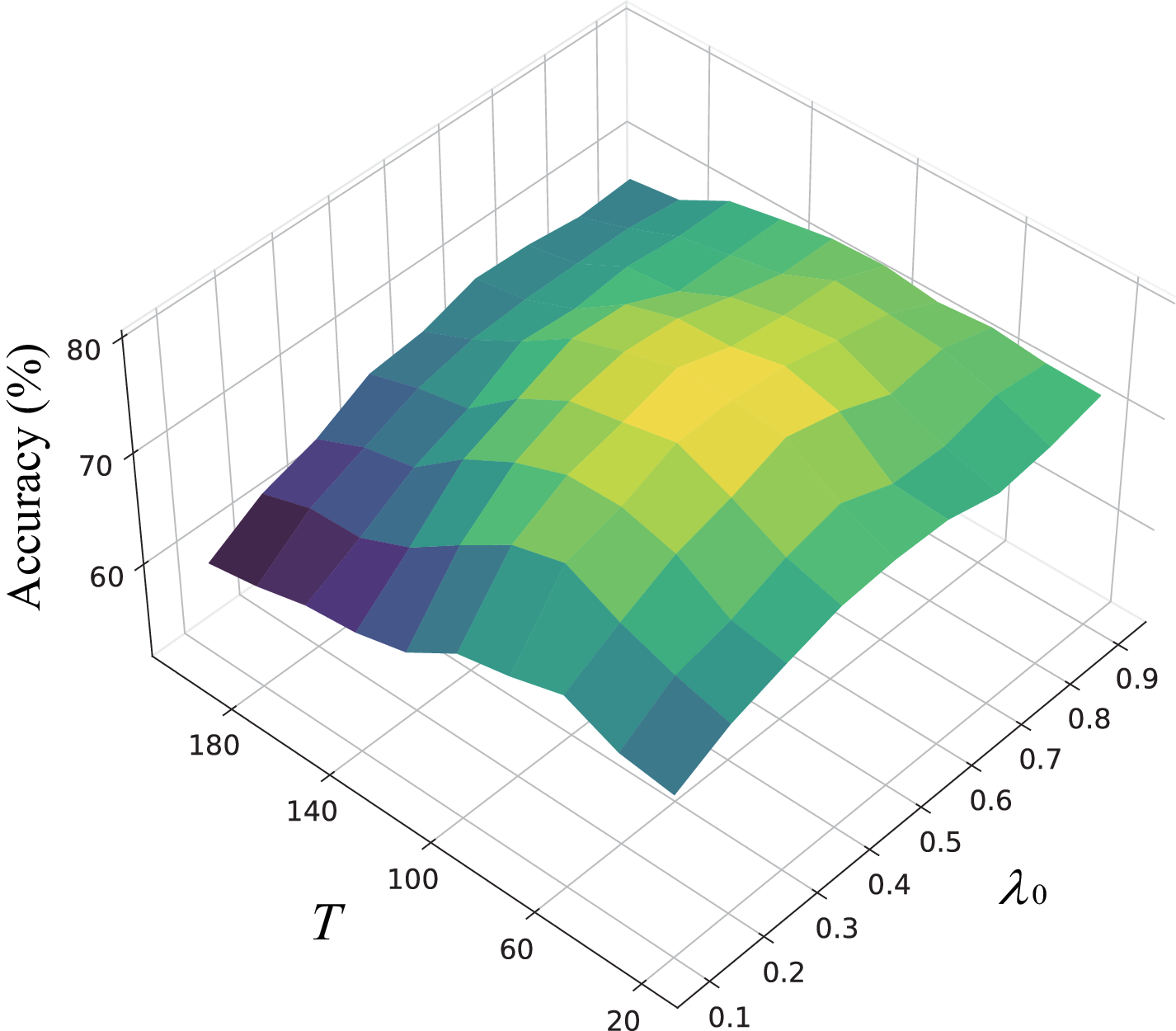}
    \caption{Parameter sensitivity analysis on Cora.}
    \label{hyper-parameter}
\end{figure} 

\subsection{Parameter Sensitivity Analysis}

Last but not least, we investigate how the hyper-parameters $\lambda_0$ and $T$ affect the performance of CLNode. $\lambda_0$ controls the initial number of training nodes, while $T$ controls the speed at which \textit{difficult nodes} are introduced to the training process. To explore the parameter sensitivity, we alter $\lambda_0$ and $T$ from \{0.1, 0.2,..., 0.9\} and \{20, 40,..., 200\}, respectively. We use GCN as the backbone GNN and report the results on Cora under random splits. The results in Figure \ref{hyper-parameter} show the following:  (1) Generally, with increasing $\lambda_0$, the performance tends to first increase and then decrease; specifically, the performance is relatively good when $\lambda_0$ is between 0.3 and 0.7.  A too small $\lambda_0$ results in few training nodes in the initial training process, meaning that the model cannot learn efficiently. In contrast, an overly large $\lambda_0$ introduces \textit{difficult nodes} during initial training and thus degrades the accuracy.  (2) Similarly, as $T$ increases, the test accuracy tends to first increase and then decrease. A too small $T$ will quickly introduce more \textit{difficult nodes}, thus degrading the backbone GNN's performance; conversely, an extremely large $T$ causes the backbone GNN to be trained mainly on the easy subset, causing a loss of the information contained in \textit{difficult nodes}.

\section{Conclusion}
In this paper, we study the problem of training GNNs on uneven-quality training nodes. Current GNNs assume that all training nodes contribute equally during training; as a result, \textit{difficult nodes} degrade their accuracy and robustness. To address these issues, we propose a novel framework CLNode to mitigate the detrimental effect of \textit{difficult nodes}. Specifically, we design a multi-perspective difficulty measurer to accurately measure node difficulty using local and global information. Based on these measurements, a continuous training scheduler is proposed to feed nodes to the training progress in an easy-to-difficult curriculum. Extensive experiments on five benchmark datasets demonstrate that CLNode is a general framework that can be combined with six representative backbone GNNs to improve their accuracy. Further experiments are conducted on noisily labeled graphs to prove that CLNode enhances backbone GNNs' robustness. An interesting future direction to expand the current work is to explore the application of curriculum learning to more graph-related tasks, e.g., link prediction.

\section{Acknowledgments}
This work was supported in part by the Natural Science Foundation of China (Nos. 61976162, 82174230), Artificial Intelligence Innovation Project of Wuhan Science and Technology Bureau (No.20\\22010702040070), Science and Technology Major Project of Hubei Province (Next Generation AI Technologies) (No. 2019AEA170), and Joint Fund for Translational Medicine and Interdisciplinary Research of Zhongnan Hospital of Wuhan University (No. ZNJC202016).


\end{document}